\documentclass{article}

     \usepackage[final, nonatbib]{neurips_2020}

\usepackage[utf8]{inputenc} 
\usepackage[T1]{fontenc}    
\usepackage{hyperref}       
\usepackage{url}            
\usepackage{booktabs}       
\usepackage{amsfonts}       
\usepackage{nicefrac}       
\usepackage{microtype}      
\usepackage{graphicx}

\title{Semantic Video Segmentation for Intracytoplasmic Sperm Injection Procedures}

\author{%
  Peter He \\
  Department of Computing\\
  Imperial College London\\
  London, UK\\
  \texttt{ph1718@ic.ac.uk} \\
  \And
  Raksha Jain \\
  Faculty of Medicine \\
  Imperial College London \\
  London, UK \\
  \AND 
  Jérôme Chambost \\
  Apricity \\
  Paris, France \\
  \And
  Céline Jacques \\
  Apricity \\
  Paris, France \\
  \And
  Cristina Hickman \\
  Institute of Reproductive and \\ Developmental Biology \\
  Imperial College London \\
  London, UK\\
}

\begin{document}

\maketitle

\begin{abstract}
We present the first deep learning model for the analysis of intracytoplasmic sperm injection (ICSI) procedures. Using a dataset of ICSI procedure videos, we train a deep neural network to segment key objects in the videos achieving a mean IoU of 0.962, and to localize the needle tip achieving a mean pixel error of 3.793 pixels at 14 FPS on a single GPU. We further analyze the variation between the dataset's human annotators and find the model's performance to be comparable to human experts.

\end{abstract}

\section{Introduction}

Intracytoplasmic sperm injection (ICSI) is an assisted reproductive technology (ART) involving the injection of a single spermatozoon directly into the cytoplasm of a mature oocyte under a microscope. Though used in 70-80\% of in-vitro fertilization (IVF) cycles, some technical aspects of the procedure remain controversial or inconclusive \cite{10.1093/humupd/dmv050}. Moreover, despite increasing standardization \cite{10.1093/humupd/dmv050}, the success rate of the procedure can vary from operator to operator \cite{TIEGS202019,daniel_hickman_wilkinson_oliana_gwinnett_trew_lavery_2015,SHEN2003355} with clinic success rates in North America ranging from 50-80\% \cite{5755184}. 

A number of studies attribute these variations (at least in part) to operator technique \cite{TIEGS202019,daniel_hickman_wilkinson_oliana_gwinnett_trew_lavery_2015,SHEN2003355, 10.1093/humrep/deh325}. A prospective study in \cite{10.1093/humrep/deh325} found that a modified injection technique led to "adequate" fertilization and pregnancy rates in patients with previous ICSI failures using the conventional technique. A retrospective study of 535 manually-analyzed videos of ICSI procedures in \cite{daniel_hickman_wilkinson_oliana_gwinnett_trew_lavery_2015} found that a certain type of intracellular needle movement can significantly reduce the likelihood of fertilization, providing a measurable technical performance indicator that could be implemented as part of an ICSI quality control process.

In this study, we propose and implement the first deep neural network model for the segmentation of key objects in ICSI procedure videos. The model has applications in not only accelerating the hitherto slow and manual task of analyzing ICSI videos for research, but also in implementing quality control processes in the IVF laboratory and providing trainee embryologists with real-time feedback on their technique.

\section{Method}
\subsection{Data Preparation}
Videos of 156 ICSI procedures were obtained from a private clinic across four embryologists on three different ICSI kits. The videos were recorded at 15 FPS and were split between training (130), validation (3) and testing (23) sets. Frames were extracted from each trimmed video once every three frames yielding a dataset of 7983 frames. The frames were labelled with polygons being drawn around the suction pipette and oolemma and a point being placed at the needle tip. Each frame was labelled by one of a team of five operators and validated by a sixth. The frames were then converted to grayscale, resized to $512 \times 512$ pixels and processed with contrasted limited adaptive histogram equalization.

\begin{figure}[!t]
\begin{center}
\centerline{\includegraphics[width=\columnwidth]{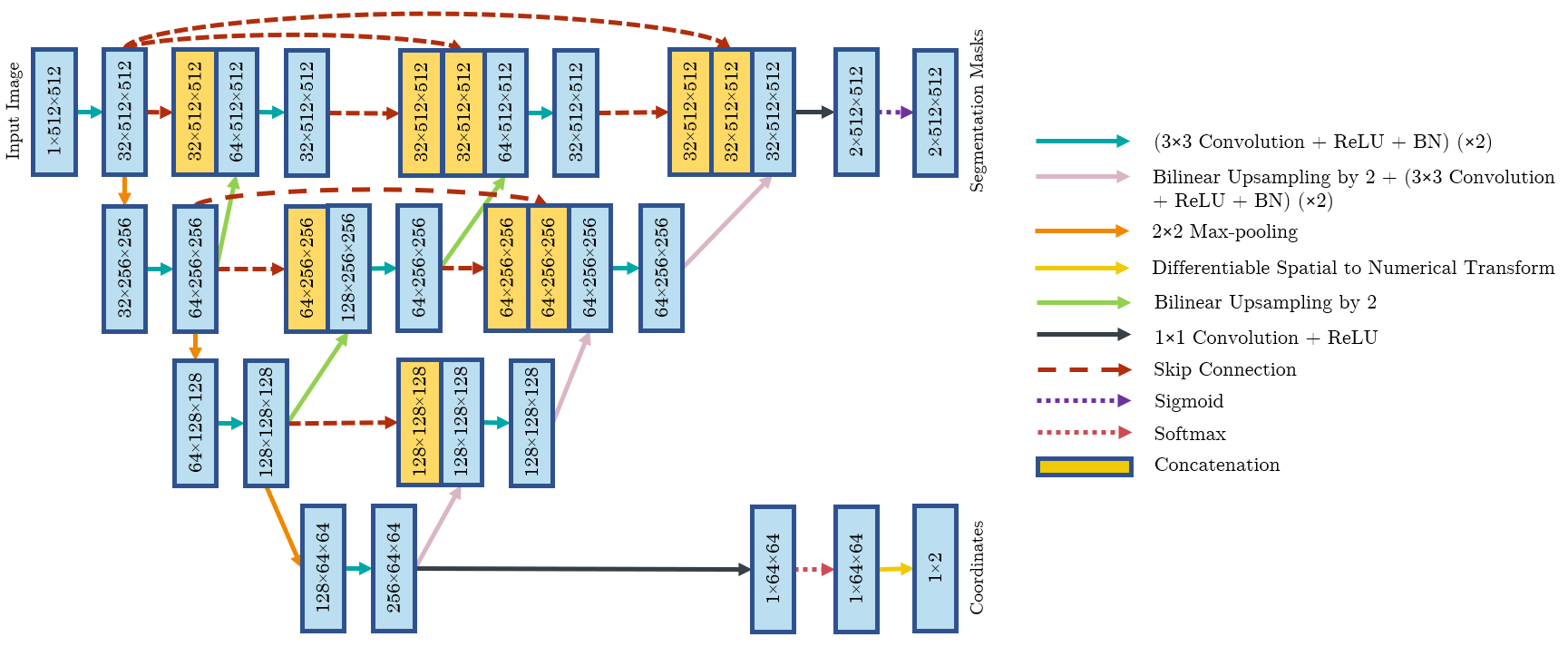}}
\caption{Model Architecture}
\label{arch}
\end{center}
\end{figure}

\subsection{Model Architecture}
We propose a modified nested U-Net architecture based on \cite{zhou_siddiquee_tajbakhsh_liang_2020} with multiple heads (as detailed in Figure \ref{arch}). The encoder takes some $512 \times 512$ input and downsamples it to a $64 \times 64 \times 256$ bottleneck through successive layers of two $3 \times 3$ convolutions followed by batch normalization and $2 \times 2$ max-pooling. From the bottleneck, the network splits to a decoder branch
and a needle localization branch. Each layer of the decoder branch comprises of bilinear upsampling followed by  two $3 \times 3$ convolutions, concatenation with nested skip pathways, a further two $3 \times 3$ convolutions and batch normalization. A final $3 \times 3$ convolution attached to the end of the decoder generates segmentation masks $y_{seg} \in [0,1]^{512 \times 512 \times 2}$. The needle localization module comprises of a $1 \times 1$ convolution followed by a softmax layer which generates a normalized heatmap $\mathit{Z} \in [0,1]^{64 \times 64}$ for the needle tip position. The normalized heatmap is passed through a differentiable spatial to numerical transform \cite{DBLP:journals/corr/abs-1801-07372} to produce a corresponding pair of coordinates $y_{coords} \in [-1,1]^{2}$.

\subsection{Model Training}
The model is trained with the multi-objective loss:

$$ \mathcal{L}_{total} = \mathcal{L}_{seg} + \lambda_{1} \mathcal{L}_{euc} + \lambda_{2} \mathcal{L}_{js} $$

where $\lambda_{1}, \lambda_{2} \in \mathbb{R}$ are constant weightings; $\mathcal{L}_{seg}$ is the Dice loss between $y_{seg}$ and the ground truth segmentation masks; $\mathcal{L}_{euc}$ is the Euclidean distance between the $y_{coords}$ and the ground truth needle tip coordinates; and $\mathcal{L}_{js}$ is a the Jensen-Shannon divergence between $\mathit{Z}$ and a normal distribution around the ground truth needle tip coordinates as described in \cite{DBLP:journals/corr/abs-1801-07372}.

The parameters of the model were learnt using the diffGrad optimizer \cite{8939562} with batches of size 4 and an initial learning rate of $1\times10^{-3}$. The training data was augmented with random cropping, rotation, flips, elastic and optical distortion, Gaussian noise and erasing. 

\section{Experiments}

\subsection{Intra \& Interoperator Variation}
In order to understand the variation in labels between labelling operators, each operator was asked to label a smaller dataset of 14 frames randomly selected from the training set. The IoU between segmentation masks as well as the Euclidean distance between the needle tip annotations were calculated for each pairing of operators.

Moreover, the variation in labels from a single operator was quantified: the operators were asked to relabel the same dataset of 14 frames an additional four times. Successive rounds of labelling were spaced at least two hours apart and annotations from previous rounds were hidden. The IoU between segmentation masks and the Euclidean distance between the needle tip annotations were calculated for each operator over each pairing of rounds. 

The results for both experiments are summarized in Table \ref{var}. It was determined that there was no statistically significant difference between intra and interoperator performance ($p > .05$).

\begin{figure}[!t]
   \begin{minipage}{0.48\columnwidth}
     \centering
     \includegraphics[width=0.9\linewidth]{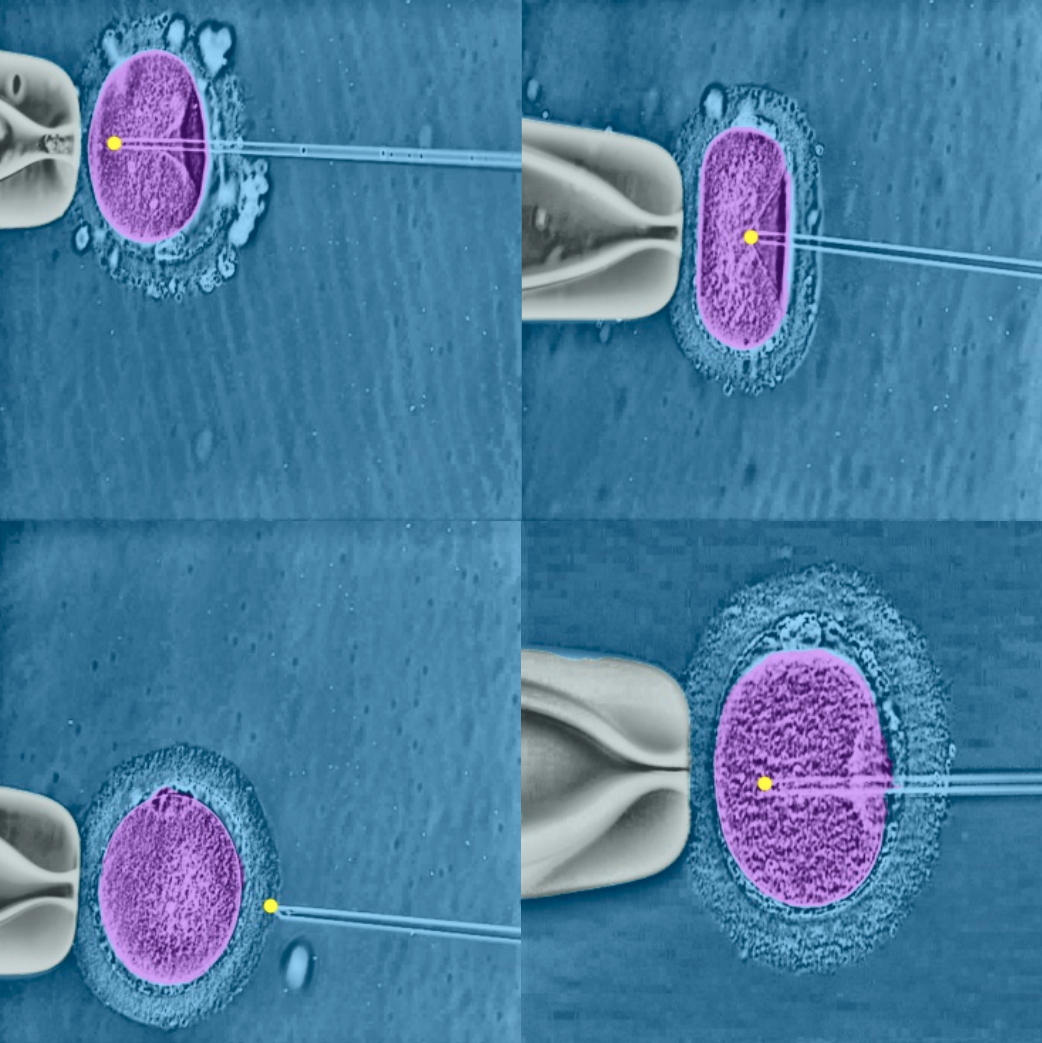}
     \caption{Examples of predicted segmentation masks. The oolemma is highlighted purple; the suction pipette is highlighted grey; the background is highlighted blue; and the predicted needle tip is shown with a yellow dot.}\label{Fig:Data1}
     \label{examples}
   \end{minipage}\hfill
   \begin{minipage}{0.48\columnwidth}
      \centering
    \includegraphics[width=\linewidth]{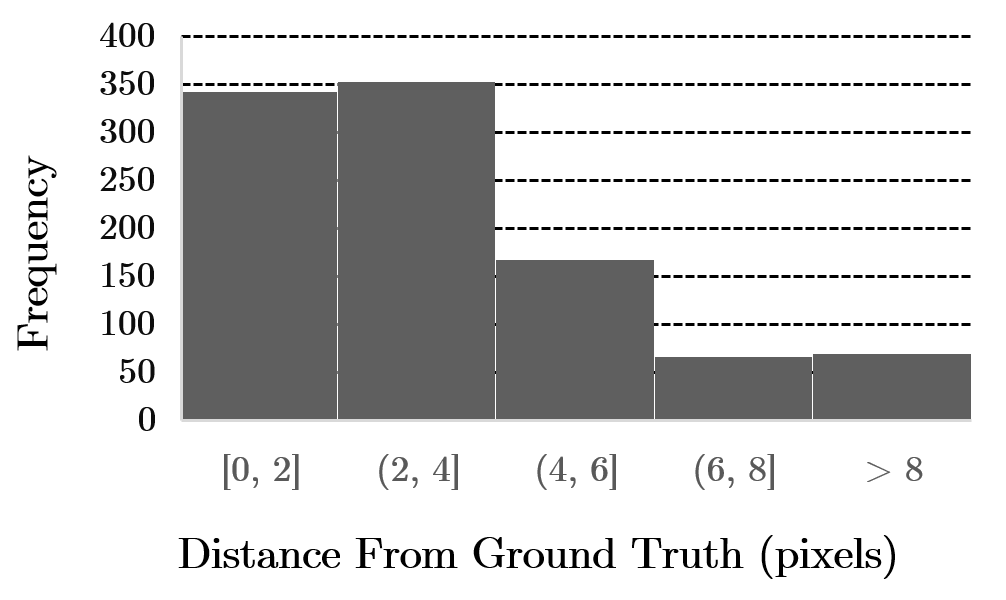}
     \caption{Histogram of distances from needle tip predictions to ground truth.}
     \label{hist}
        \renewcommand{\figurename}{Table}
   \setcounter{figure}{0} 
      \caption{Mean Operator Performance (with standard deviations in square brackets).}
      \label{var}
      \begin{tabular}{lll}
        \toprule
             & Interoperator     & Intraoperator \\
        \midrule
        Oolemma (IoU) & 0.960 [0.016] & 0.959 [0.015]      \\
        Pipette (IoU) & 0.964 [0.021] & 0.965 [0.013]     \\
        Needle (pixels) & 3.731 [3.043] & 3.503 [3.230]  \\
        \bottomrule
      \end{tabular}

     \end{minipage}
\end{figure}

\subsection{Model Evaluation}
The model was trained for 31 epochs (26 hours) on a single RTX 2080Ti GPU and evaluated on 1000 frames extracted from 23 ICSI videos. The model achieved IoU scores of 0.961 ($\sigma$ = 0.021) and 0.963 ($\sigma$ = 0.064) for the oolemma and pipette classes respectively. The average Euclidean distance from the predicted needle tip to the ground truth location was 3.793 ($\sigma$ = 6.981). A histogram of these distances can be seen in Figure \ref{hist}. It was determined that there was no significant difference between human interoperator and our model's performance on any of the classes ($p > .05$). The model is relatively small at 2.6 M parameters and inference time was 70.2 ms per frame (\textasciitilde 14 FPS).

\section{Conclusion \& Future Work}
In this paper, we introduce the first deep neural network model for the segmentation of ICSI videos. The model achieves roughly human-level performance with speed and consistency on a single GPU setting a strong baseline for future research in the area. 

There is much scope for further work on the model itself. Temporal consistency between predictions for consecutive frames may be improved by conditioning predictions on previous frames or through post-processing techniques (though the latter may not always be possible in real-time settings). Though already suitable for the analysis of ICSI technique, the model may prove useful to a wider audience by recognizing other structures present in the videos. Moreover, in order to increase robustness with respect to different laboratory setups, the training dataset should be expanded to encompass a wider range of equipment across multiple clinics. At a more applied level, the model may be combined with unsupervised techniques to analyze ICSI technique at scale and aid in the discovery of "best practices" in ICSI.

\section*{Broader Impact}

This work has the potential to positively impact infertile people and others unable to conceive without medical assistance by enabling further research into ICSI which may help them in fulfilling their dreams of having children. However, the research may also generate a negative impact by perhaps enabling further work into automated ICSI procedures which, while obviously having applications in improving efficiency and standardization in IVF laboratories, may be misused in the long run (for example, for eugenics or the raising of an army of clones). Moreover, in the shorter term, errors made by the system when used for processing videos for research may lead to incorrect conclusions being drawn. It is thus important that predictions are validated by a human-in-the-loop and not blindly trusted.

\begin{ack}
The project would not have been possible without the participation of labelling operators \textbf{Alyssa Arshad} (Faculty of Life Sciences and Medicine, King's College London), \textbf{Ryan Patel} (Barts and the London School of Medicine and Dentistry, Queen Mary University of London), \textbf{Sahar Ley} (School of Biological and Chemical Sciences, Queen Mary University of London) and \textbf{Urvi Bihani} (Faculty of Medicine, Imperial College London). We also thank \textbf{Bogdan Surdu} for his comments and proofreading.

Funding in direct support of this work: none.  
\end{ack}

\bibliographystyle{unsrt}
\bibliography{bib}

\begin{thebibliography}{1}

\bibitem{10.1093/humupd/dmv050}
Patrizia Rubino, Paola Viganò, Alice Luddi, and Paola Piomboni.
\newblock {The ICSI procedure from past to future: a systematic review of the
  more controversial aspects}.
\newblock {\em Human Reproduction Update}, 22(2):194--227, 11 2015.

\bibitem{TIEGS202019}
Ashley~W Tiegs and Richard~T Scott.
\newblock Evaluation of fertilization, usable blastocyst development and
  sustained implantation rates according to intracytoplasmic sperm injection
  operator experience.
\newblock {\em Reproductive BioMedicine Online}, 41(1):19 -- 27, 2020.

\bibitem{daniel_hickman_wilkinson_oliana_gwinnett_trew_lavery_2015}
C.E. Daniel, C.~Hickman, T.~Wilkinson, O.~Oliana, D.~Gwinnett, G.~Trew, and
  S.~Lavery.
\newblock {Maximising success rates by improving ICSI technique: which factors
  affect outcome?}
\newblock {\em Fertility and Sterility}, 104(3), 2015.

\bibitem{SHEN2003355}
Shehua Shen, Amin Khabani, Nancy Klein, and David Battaglia.
\newblock Statistical analysis of factors affecting fertilization rates and
  clinical outcome associated with intracytoplasmic sperm injection.
\newblock {\em Fertility and Sterility}, 79(2):355 -- 360, 2003.

\bibitem{5755184}
Z.~{Lu}, X.~{Zhang}, C.~{Leung}, N.~{Esfandiari}, R.~F. {Casper}, and Y.~{Sun}.
\newblock {Robotic ICSI (Intracytoplasmic Sperm Injection)}.
\newblock {\em IEEE Transactions on Biomedical Engineering}, 58(7):2102--2108,
  2011.

\bibitem{10.1093/humrep/deh325}
T.~Ebner, M.~Moser, M.~Sommergruber, K.~Jesacher, and G.~Tews.
\newblock {Complete oocyte activation failure after ICSI can be overcome by a
  modified injection technique}.
\newblock {\em Human Reproduction}, 19(8):1837--1841, 08 2004.

\bibitem{zhou_siddiquee_tajbakhsh_liang_2020}
Zongwei Zhou, Md~Mahfuzur~Rahman Siddiquee, Nima Tajbakhsh, and Jianming Liang.
\newblock {UNet++: Redesigning Skip Connections to Exploit Multiscale Features
  in Image Segmentation}.
\newblock {\em IEEE Transactions on Medical Imaging}, 39(6):1856--1867, 2020.

\bibitem{DBLP:journals/corr/abs-1801-07372}
Aiden Nibali, Zhen He, Stuart Morgan, and Luke Prendergast.
\newblock {Numerical Coordinate Regression with Convolutional Neural Networks}.
\newblock {\em CoRR}, abs/1801.07372, 2018.

\bibitem{8939562}
S.~R. {Dubey}, S.~{Chakraborty}, S.~K. {Roy}, S.~{Mukherjee}, S.~K. {Singh},
  and B.~B. {Chaudhuri}.
\newblock diffgrad: An optimization method for convolutional neural networks.
\newblock {\em IEEE Transactions on Neural Networks and Learning Systems},
  pages 1--12, 2019.

\end{thebibliography}

\end{document}